\documentclass[letterpaper]{article} 
\usepackage{aaai2026}  
\usepackage{times}  
\usepackage{helvet}  
\usepackage{courier}  
\usepackage[hyphens]{url}  
\usepackage{graphicx} 
\usepackage{natbib}  
\usepackage{caption} 
\usepackage{algorithm}
\usepackage{algorithmic}
\usepackage{listings}
\DeclareCaptionStyle{ruled}{labelfont=normalfont,labelsep=colon,strut=off} 

\floatstyle{ruled}
\newfloat{listing}{tb}{lst}{}
\floatname{listing}{Listing}
\usepackage{booktabs}
\usepackage{array} 
\usepackage{multirow}

\title{Intelligent Human-Machine Partnership for Manufacturing: Enhancing Warehouse Planning through Simulation-Driven Knowledge Graphs and LLM Collaboration}
\author {
    Himabindu Thogaru\textsuperscript{\rm 1},
Saisubramaniam Gopalakrishnan\textsuperscript{\rm 1},
 Zishan Ahmad\textsuperscript{\rm 1}, Anirudh Deodhar\textsuperscript{\rm 1} \thanks{Corresponding author. }
}
\affiliations {
    \textsuperscript{\rm 1}Phi Labs, Quantiphi\\

    himabindu.thogaru@quantiphi.com, saisubramaniam.gopalakrishnan@quantiphi.com,\\ zishan.ahmad@quantiphi.com,
    anirudh.deodhar@quantiphi.com
    
}
\copyrighttext{Accepted at the AAAI 2026 Workshop on Human-Centric Manufacturing, Singapore.}

\begin{document}

\maketitle

\begin{abstract}
 Manufacturing planners face complex operational challenges that require seamless collaboration between human expertise and intelligent systems to achieve optimal performance in modern production environments. Traditional approaches to analyzing simulation-based manufacturing data often create barriers between human decision-makers and critical operational insights, limiting effective partnership in manufacturing planning. Our framework establishes a collaborative intelligence system integrating Knowledge Graphs and Large Language Model-based agents to bridge this gap, empowering manufacturing professionals through natural language interfaces for complex operational analysis. The system transforms simulation data into semantically rich representations, enabling planners to interact naturally with operational insights without specialized expertise. A collaborative LLM agent works alongside human decision-makers, employing iterative reasoning that mirrors human analytical thinking while generating precise queries for knowledge extraction and providing transparent validation. This partnership approach to manufacturing bottleneck identification, validated through operational scenarios, demonstrates enhanced performance while maintaining human oversight and decision authority. For operational inquiries, the system achieves near-perfect accuracy through natural language interaction. For investigative scenarios requiring collaborative analysis, we demonstrate the framework's effectiveness in supporting human experts to uncover interconnected operational issues that enhance understanding and decision-making. This work advances collaborative manufacturing by creating intuitive methods for actionable insights, reducing cognitive load while amplifying human analytical capabilities in evolving manufacturing ecosystems.
\end{abstract}

\section{Introduction}
Modern manufacturing systems represent complex ecosystems where human expertise and advanced technology must work in seamless partnership to achieve optimal performance. As Industry 4.0 and smart manufacturing initiatives transform production environments, the challenge of effective human-machine collaboration has become paramount \cite{ZHONG2017616, GIUSEPPE}. Human decision-makers in manufacturing planning bring irreplaceable domain knowledge, contextual understanding and adaptive problem-solving capabilities, while AI systems offer computational power, pattern recognition and data processing at scale. However, realizing the full potential of this partnership requires bridging the gap between human cognitive processes and complex technological systems, particularly in the analysis of intricate operational data. \\
Manufacturing facilities, including warehouses, distribution centers and production floors, are characterized by sophisticated interactions between personnel, automated equipment, processes and physical layouts \cite{DEKOSTER2007481, LEEC,GUGOET} . The complexity of these human-machine collaborative environments demands decision support systems that can interpret vast amounts of operational data while remaining accessible and interpretable to human planners. Discrete Event Simulation (DES) has emerged as a fundamental tool for modeling these systems \cite{banks2005discrete, law2000simulation}, enabling stakeholders to evaluate performance, test design alternatives and understand system dynamics. However, a critical challenge persists: how can human decision-makers effectively collaborate with AI systems to extract actionable insights from the voluminous and highly granular output data that DES generates? \\
Traditional approaches to simulation analysis often create barriers between human expertise and technological capabilities. Conventional methods, frequently reliant on manual inspection of aggregate statistics \cite{law2000simulation} or development of custom scripts tailored to specific simulation outputs, are not only time-intensive and error-prone but also limit the ability of human planners to engage in meaningful collaboration with AI systems. These approaches fail to leverage the complementary strengths of human intuition and machine computation, often requiring specialized technical expertise that creates silos between operational planners and analytical tools. As manufacturing environments become increasingly complex and dynamic, there is an urgent need for human-centric AI systems that can facilitate effective collaboration between human decision-makers and advanced analytical capabilities.\\
The emergence of Artificial Intelligence (AI) in manufacturing and logistics \cite{ivanov2019impact,ZHONG2017616, app13116746}  presents unprecedented opportunities to create collaborative intelligence systems that augment rather than replace human expertise. However, realizing these opportunities requires addressing fundamental challenges in human-AI interaction: How can AI systems interpret user intentions while managing procedural uncertainties? How can humans and machines engage in bidirectional learning? How can collaborative partnerships be designed to achieve mutually beneficial goals in evolving manufacturing ecosystems?\\
To address these human-centric manufacturing challenges, our work proposes a novel framework that integrates Knowledge Graphs (KGs) \cite{hogan2021knowledge} and Large Language Model (LLM)s \cite{zhao2023survey, pan2023large} to enable effective human-AI collaboration in manufacturing planning through natural language interfaces for complex operational analysis. Our approach recognizes that the core challenge is not merely technical analysis, but rather creating systems that facilitate meaningful partnership between human planners and AI capabilities.\\
The foundation of our human-centric approach lies in transforming complex data generated by DES into semantically rich Knowledge Graphs that both humans and AI systems can effectively utilize. By representing simulation output as graphs, we enable the intricate dependencies and flows within manufacturing systems to be explicitly captured and collaboratively explored by human planners and AI agents. While KGs are increasingly applied to analyze real-world industrial and supply chain data for enhanced visibility and risk management \cite{noy2019industry,kosasih2024towards}, their application to creating human-AI collaborative interfaces for simulation analysis remains relatively unexplored.\\
Building upon this structured representation, our framework employs LLM-based agents to create intuitive natural language interfaces that allow human planners to engage in collaborative analysis without requiring specialized technical expertise. This approach democratizes access to complex analytical capabilities, enabling operations analysts, industrial engineers and manufacturing planners to engage in meaningful dialogue with AI systems using natural language queries. The LLM agent serves not as a replacement for human expertise, but as an intelligent partner that can translate human intentions into precise analytical operations while maintaining transparency and interpretability.\\
Our collaborative framework features an iterative reasoning mechanism \cite{luo2023reasoning} that enables bidirectional learning between human planners and AI systems. Given complex queries from human users, the AI agent autonomously generates sequences of sub-questions, each informed by both accumulated evidence and potential human feedback. This creates opportunities for human experts to guide the analytical process, validate intermediate findings and contribute domain knowledge that enhances the quality of insights. The agent's ability to perform self-reflection \cite{huang2022large, madaan2023self} and correct its analytical pathway creates a transparent collaborative process where humans can understand, validate and improve AI reasoning.\\
This synthesis moves beyond traditional automation paradigms to create collaborative intelligence that leverages the unique strengths of both human expertise and AI capabilities. Human planners bring contextual understanding, strategic thinking and adaptive problem-solving, while AI systems contribute computational power, pattern recognition and systematic analysis. The result is a manufacturing planning assistant that facilitates reciprocal partnership, where humans and machines co-develop solutions, engage in mutual learning and achieve shared goals in dynamic manufacturing environments.

\begin{figure}
  \centering

  \includegraphics[width=0.9\linewidth]{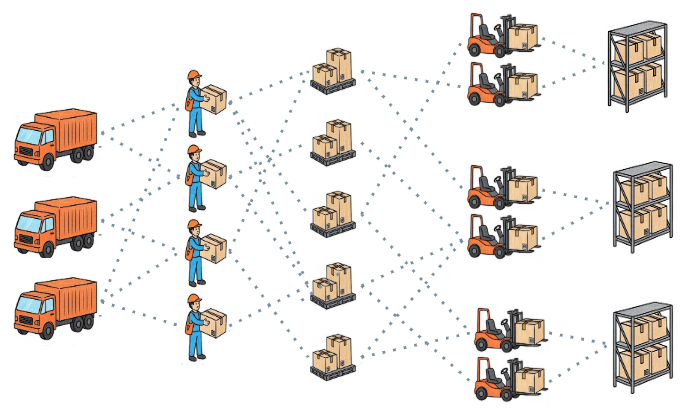}
  \small
  \caption{Workflow diagram of the warehouse unloading process, detailing the sequence from supplier deliveries and worker handling, through stages involving Automated Guided Vehicles (AGVs) and forklifts, to packages formed for storage. Numerical figures are for representational purposes only, the exact numbers are provided in text.}
  \label{fig:simulation_flow}
\end{figure}

  \vspace{-1mm}

\section{Related Work}

\subsection{Manufacturing Intelligence and Simulation Analysis}
Modern manufacturing systems require sophisticated analytical approaches to manage complex interactions between personnel, equipment and processes \cite{zhong2017intelligent}. DES has become a fundamental tool for modeling these systems, enabling stakeholders to evaluate performance and test design alternatives \cite{banks2005discrete,law2000simulation}. In the Industry 4.0 context, DES serves as a core component of Digital Twins, providing real-time operational insights for manufacturing decision-making \cite{rasheed2020digital,leng2019digital} .\\
Despite its analytical power, extracting actionable insights from voluminous DES output data remains challenging. Traditional approaches rely on manual inspection of aggregate statistics or custom analytical scripts, which are time-intensive and often fail to uncover complex system behaviors \cite{NEGAHBAN2014241}. This has created demand for AI-driven approaches that can unlock deeper insights from simulation data while remaining accessible to manufacturing professionals.
  \vspace{-1mm}

\subsection{Knowledge Graphs in Industrial Applications}
Knowledge Graphs have gained prominence as powerful tools for representing and reasoning over complex industrial data \cite{noy2019industry}. In manufacturing contexts, KGs enable structured representation of operational relationships, supporting applications in supply chain visibility, risk management and inventory optimization \cite{saidi2025modeling}. Also this work has applied KGs to warehouse robot operations \cite{kattepur2019roboplanner} and production logistics \cite{zhao2022digital}, demonstrating their potential for operational analysis.\\
However, the application of KGs to structure and analyze DES output data remains relatively unexplored. While KGs excel at representing complex relational data, their specific use for simulation-based manufacturing planning represents a significant opportunity for enhancing analytical capabilities.
  \vspace{-1mm}

\subsection{LLMs and Knowledge Graph Integration}
The integration of LLMs with Knowledge Graphs has emerged as a powerful approach for creating accessible AI systems \cite{pan2023large, zhao2023survey}. This integration addresses key challenges: KGs ground LLMs with structured facts, reducing hallucinations and improving reliability \cite{agrawal2023can}, while LLMs make KG information accessible through natural language interfaces, eliminating the need for specialized query languages.\\
Several integration patterns have emerged. KG-enhanced LLMs leverage structured knowledge during inference, often through Retrieval-Augmented Generation approaches. LLM-augmented KGs use language models for construction and enrichment. Synergized approaches feature deeper integration with LLM-based agents that reason over and interact with KGs for complex, multi-step tasks \cite{jiang2024kg,luo2023reasoning}.
  \vspace{-1mm}

\subsection{Industrial Applications of KG-LLM Systems}
Recent developments in manufacturing include enhanced querying capabilities for accessing operational knowledge through natural language \cite{hovcevar2024integrating}, AI-driven logistics optimization \cite{ieva2025enhancing} and domain-specific question answering systems \cite{li2024knowledge}. Frameworks like SparqLLM \cite{arazzi2025augmented} have improved KG querying reliability for industrial applications.\\
Despite these advances, the application of KG-LLM systems to analyzing DES output data remains largely unexplored. Critical research questions include: how effectively can LLM-based agents transform natural language queries about simulated performance into precise KG queries, iteratively refine analytical pathways based on retrieved evidence and synthesize disparate information to diagnose operational issues with explainable reasoning.\\
Our work addresses this gap by proposing a novel framework that integrates KGs and LLM-based agents for iterative analysis of DES data, specifically targeting bottleneck identification and root cause analysis in manufacturing operations. This approach bridges traditional simulation modeling with modern AI capabilities, providing an intuitive interface for complex operational analysis.

  \vspace{-2mm}

\section{Methodology}
We implement a comprehensive two-stage framework that transforms raw DES output data into actionable warehouse planning insights through Knowledge Graph construction and Large Language Model-based reasoning. This framework addresses the critical challenge of analyzing voluminous simulation data by creating a semantically rich representation that enables natural language querying and sophisticated diagnostic analysis. The methodology bridges traditional simulation modeling with modern AI capabilities, providing warehouse planners with an intuitive interface for complex operational analysis.

\subsection{Stage 1: Comprehensive Simulation Environment Setup}
Our evaluation foundation consists of a detailed DES model representing a warehouse facility engaged in unloading, internal transport and storage operations. The simulation replicates real-world logistics environments with high fidelity, capturing the complex interactions between personnel, equipment and operational processes that characterize modern warehouse operations.

The simulated warehouse environment incorporates five distinct resource categories operating in coordinated fashion. External suppliers arrive as trucks carrying package shipments, with each supplier transporting between 30-35 packages sampled from a uniform distribution. These suppliers operate at 20 km/hr movement speed, with the facility supporting a maximum of three simultaneous unloading operations to reflect real-world dock capacity constraints.

Warehouse personnel consist of twelve employees organized into specialized teams of four workers each, with exclusive assignment to individual suppliers during unloading operations. Workers operate at 2 km/hr movement speed and handle single packages, reflecting standard warehouse safety and efficiency protocols. This team-based structure ensures systematic coordination while maintaining operational flexibility.

The automated transportation infrastructure includes twenty Automated Guided Vehicles (AGVs) operating at 3.5 km/hr with dynamic dispatch capabilities. AGVs follow First-In-First-Out scheduling protocols and traverse an average distance of 140 meters per transport operation. This reflects the sophisticated material handling automation increasingly common in modern warehouse facilities.

Five forklift units provide vertical storage capabilities, operating at 5 km/hr with block-specific assignments. Each forklift follows FIFO package handling protocols and incorporates stochastic storage operations ranging from 60-90 seconds to represent real-world variability in storage placement activities. The storage infrastructure consists of five blocks, each containing fifteen bays with three shelves, providing total capacity for 225 packages across the facility.

The operational process follows a structured four-stage flow architecture as shown in Figure \ref{fig:simulation_flow}. Initially, supplier trucks proceed to parking areas upon arrival, awaiting dock assignment based on availability before moving to designated unloading positions. Worker teams then transfer packages from suppliers to predetermined waiting points, with handling times determined by distance calculations and walking speed parameters. Subsequently, AGVs transport packages to appropriate block-specific pickup points, with travel times calculated based on distance and vehicle speed specifications. Finally, block-dedicated forklifts collect packages and store them in available bays, incorporating both calculated travel time and stochastic storage duration components.

The simulation captures comprehensive operational data including process-specific timestamps for arrival, initiation and completion events across all stages. Equipment utilization metrics and waiting times are recorded for each resource type, while package-level tracking maintains unique identifiers throughout the entire process flow. Resource state transitions and queue statistics provide additional analytical depth for performance assessment and bottleneck identification.
\begin{figure}
  \centering
  \includegraphics[width=0.85\linewidth]{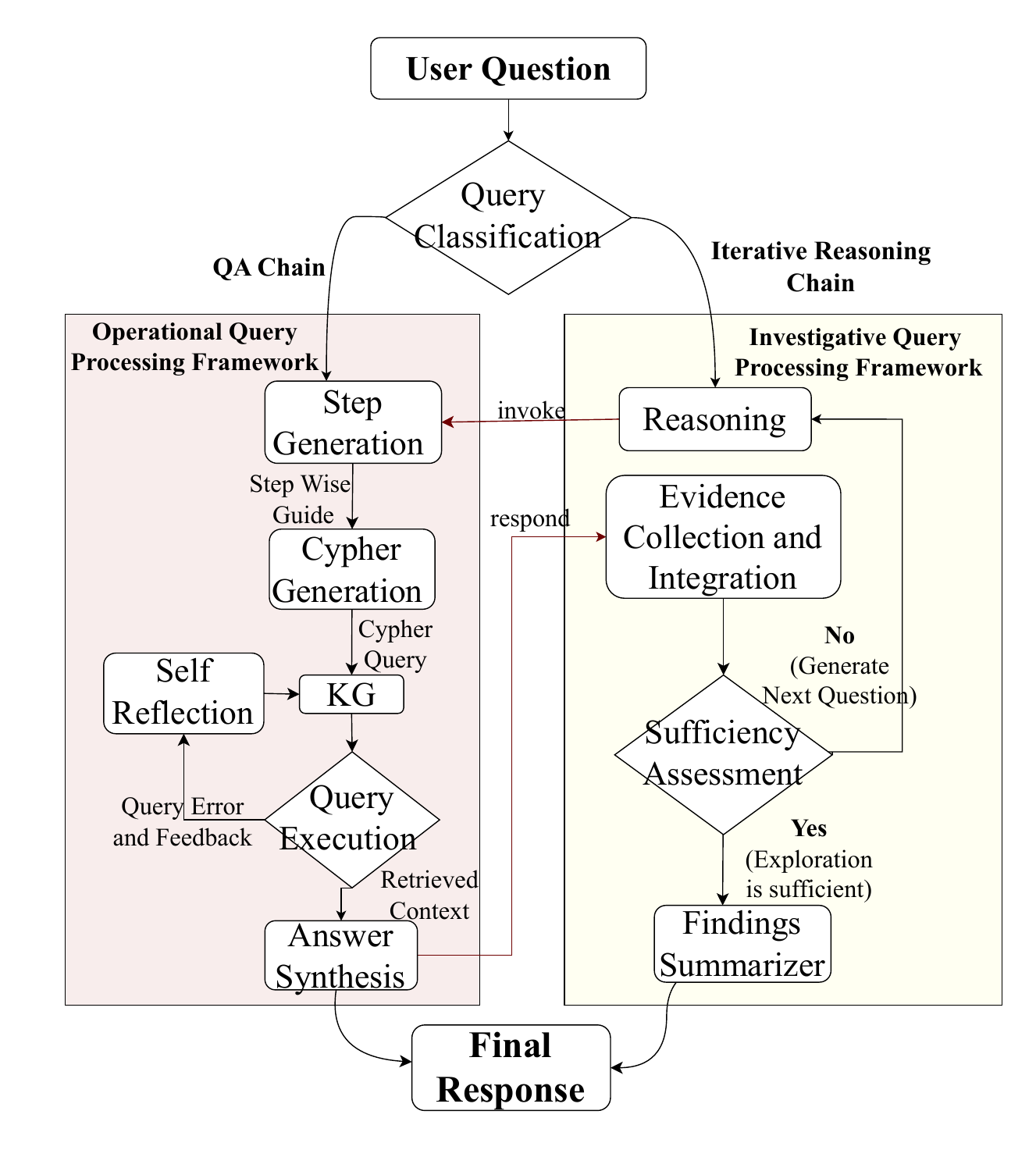}
  \small
  \caption{System architecture of the LLM Reasoning Agent comprising two components: the QA Chain and the Reasoning Chain. Queries are first classified as normal or bottleneck. Normal queries follow step-wise guidance for Cypher generation, execution and self-reflection. Bottleneck queries invoke iterative reasoning, where the agent decomposes the problem into sub-questions, gathers intermediate evidence and dynamically refines its analysis toward a final answer.}
  \label{fig:qa_architecture_agent}
\end{figure}
\subsection{Knowledge Graph Schema Design and Construction}
We developed a specialized ontology tailored specifically for warehouse simulation data representation, capturing both static resource configurations and dynamic package flow relationships. The schema design prioritizes semantic richness while maintaining query efficiency for complex analytical operations. More information about Node schema and Relationship schema is mentioned in the appendix \ref{appendix:KGSchema}.

The automated Knowledge Graph construction pipeline transforms DES output logs through a systematic five-step process. Initially, simulation event logs undergo parsing and temporal data extraction, ensuring accurate timestamp representation across all operational events. Resource nodes are then created with their associated properties, establishing the foundational entity structure. Package flow relationships are established with timestamp annotations, creating the dynamic process flow representation. Data integrity and relationship consistency undergo comprehensive validation to ensure analytical reliability. Finally, graph structure optimization occurs for efficient Cypher querying, including appropriate indexing and relationship organization for performance enhancement.

\subsection{Stage 2: 
Dual-Path Query Processing System}
Our framework implements a sophisticated dual-path architecture initiated by automated query classification to address both direct operational inquiries and complex investigative diagnostic scenarios as shown in Figure \ref{fig:qa_architecture_agent}. This architectural approach recognizes the fundamental difference between information retrieval tasks and analytical reasoning requirements in warehouse planning contexts.

The query classification module employs natural language processing to automatically categorize incoming queries into two distinct pathways. Operational queries represent direct information retrieval scenarios requiring structured data access and straightforward analytical operations. These queries typically seek specific performance metrics, resource utilization statistics or process timing information. Investigative queries encompass complex diagnostic scenarios requiring iterative reasoning, hypothesis testing and systematic bottleneck identification through multi-stage analytical processes.

\subsection{Operational Query Processing Framework}
For operational queries, the framework implements a structured QA Chain comprising four specialized processing modules that work in coordinated fashion to ensure accurate and comprehensive responses.

The Step Generation Module decomposes complex natural language queries into structured analytical steps, recognizing that warehouse operational questions often contain multiple information requirements. This module breaks down multi-faceted questions into targeted sub-queries, with each step designed for single, focused Cypher query execution. This decomposition strategy significantly improves query success rates by reducing complexity and enabling systematic error detection and correction.

The Cypher Generation Module translates structured analytical steps into precise graph database queries, leveraging the native graph structure for expressive relationship traversal operations. This approach avoids complex SQL-style joins through graph-native query patterns that more naturally represent warehouse operational flows. The module incorporates domain-specific query templates optimized for warehouse data patterns while maintaining flexibility for novel query structures.

The Query Execution and Self Reflection Module implements robust error-handling loops for query validation and automatic correction. This module provides automatic syntax correction and retry mechanisms, ensuring query execution reliability through self-correction capabilities. The error handling includes both syntactic correction for malformed queries and semantic validation for logically inconsistent operations, significantly improving overall system reliability.

The Answer Synthesis Module processes query results beyond simple data presentation, interpreting patterns and synthesizing coherent responses that provide meaningful insights for warehouse planning decisions. This module aggregates multiple query results for comprehensive answers and contextualizes numerical findings within operational frameworks that facilitate practical decision-making.

\subsection{Investigative Query Processing Framework}
For complex investigative scenarios, the framework activates an Iterative Reasoning Chain that enables systematic diagnostic analysis through evidence-based hypothesis testing and refinement.

The Reasoning Module decomposes main diagnostic problems into sequential sub-questions, generating context-aware inquiries based on accumulated evidence from previous analytical steps. This module dynamically refines the analytical pathway through iterative questioning, ensuring that investigation depth and breadth adapt to the specific characteristics of each diagnostic scenario. The sub-question generation incorporates domain knowledge about warehouse operations to ensure investigative relevance and efficiency.

Evidence Collection and Integration operates through the complete QA Chain infrastructure for each generated sub-question, enabling focused evidence gathering through targeted Cypher queries. This approach builds comprehensive evidence bases through iterative information accumulation while maintaining analytical coherence across investigation stages. The integration process includes temporal correlation analysis and causal relationship identification to support robust diagnostic conclusions.

Self-Reflection and Validation mechanisms implement continuous validation of analytical findings throughout the investigation process. These mechanisms correct reasoning errors through self-reflection procedures and ensure diagnostic accuracy through multi-stage verification processes. The validation includes both logical consistency checking and factual accuracy verification against the knowledge graph data.

Sufficiency Assessment determines when adequate evidence has been gathered to support reliable diagnostic conclusions. This assessment triggers final synthesis when investigation completeness criteria are met while preventing over-analysis that might introduce unnecessary complexity or confusion. The sufficiency criteria incorporate both breadth of coverage across relevant operational dimensions and depth of analysis for identified bottleneck areas.

\subsection{Technical Implementation Framework}The technical implementation leverages state-of-the-art technologies and frameworks to ensure robust, scalable and reliable operation across diverse warehouse planning scenarios.

LLM Integration utilizes OpenAI's GPT-5 \cite{OpenAI_GPT5} through the LangGraph framework, with each processing module implemented through independent Large Language Model calls. This modular architecture enables specialized function optimization while maintaining system coherence. Temperature ranging from 0.0 to 0.3 ensures investigative accuracy for diagnostic scenarios while allowing appropriate creativity for complex reasoning tasks. Robust error handling and retry mechanisms provide system reliability even under challenging operational conditions.

Graph Database Integration through Neo4J \cite{neo4j_website} provides efficient Cypher query execution capabilities optimized for warehouse operational patterns. The integration supports scalable graph traversal operations that maintain performance even with large-scale simulation datasets. Optimized relationship-based data access patterns ensure rapid response times for both simple operational queries and complex investigative analyses.

Adaptive Reasoning Parameters enable dynamic question generation based on intermediate findings, ensuring that investigation pathways remain relevant and efficient as evidence accumulates. Context-aware query refinement adjusts analytical approaches based on evolving understanding of operational scenarios. Evidence-driven investigation pathway adjustment ensures that diagnostic processes adapt to the specific characteristics and requirements of each bottleneck scenario.

The integration of structured simulation data with advanced AI reasoning capabilities creates a powerful framework for transforming complex operational data into actionable warehouse planning insights that support both tactical and strategic decision-making processes.
\section{Warehouse Operations Scenario Design}

\subsection{Simulation Set-up}

This study is based on the data generated by an in-house DES model that includes operations of a warehouse facility engaged in the unloading, internal transport and storage of incoming packages. The simulation is designed to replicate real-world warehouse logistics, capturing the interactions between key resources such as suppliers (trucks), workers, automated guided vehicles (AGVs), forklifts and storage infrastructure. The details of the simulation including equipment-resource specifications, process flow,  operational assumptions and data captured can be found in appendix \ref{appendix:warehouse_details}. 

\subsection{Analytical Questions Set-up}
Based on data generated from the DES operating scenario, two types of questions were formulated for the evaluation of analytical capabilities of our framework as following:
\begin{table*}[!t]
\centering 
\small
\setlength{\tabcolsep}{3pt} 

\begin{tabular}{@{}c | cc | cc | cc | cc | cc | cc@{}}
\toprule
{\textbf{Method}} & \multicolumn{2}{c}{\textbf{Supplier}} & \multicolumn{2}{c}{\textbf{Worker}} & \multicolumn{2}{c}{\textbf{AGV}} & \multicolumn{2}{c}{\textbf{Forklift}} & \multicolumn{2}{c}{\textbf{Package}} & \multicolumn{2}{c}{\textbf{Average}} \\
\cmidrule(lr){2-3} \cmidrule(lr){4-5} \cmidrule(lr){6-7} \cmidrule(lr){8-9} \cmidrule(lr){10-11} \cmidrule(lr){12-13}

 & \textbf{P@1} & \textbf{P@2} & \textbf{P@1} & \textbf{P@2} & \textbf{P@1} & \textbf{P@2} & \textbf{P@1} & \textbf{P@2} & \textbf{P@1} & \textbf{P@2} & \textbf{P@1} & \textbf{P@2}\\
\midrule

Baseline    & 0.4 &	0.5 &	0.33	& 0.33	& 0.67	&0.83 &	0.9 &	1 &	0.65 &	0.9 &	0.59&	0.71
 \\
Baseline+ SR & 0.7 &	0.8 &	0.42&	0.5	& 0.75	& 1 &	0.95&	1	&0.7&	0.8&	0.70 &	0.82
\\

Our framework  & \textbf{1.00} & \textbf{1.00} & \textbf{0.75} & \textbf{1.00 }& \textbf{0.92} & \textbf{1.00} & \textbf{1.00} & \textbf{ 1.00} & \textbf{0.95} &\textbf{ 1.00} & \textbf{0.92} &\textbf{ 1.00} \\
\bottomrule
\end{tabular}
\small
\caption{Performance on Operational QA by Method and Stage (Pass@k Scores). Baseline: Single-pass Cypher query generation followed by answer synthesis. SR: Self-Reflection. Our Framework (Guided Iterative Steps): Question decomposition for structured step generation; each step involves (Cypher query + Answer Generation + Self-Reflection).}

\label{tab:operational_qa_results}
\end{table*}

\textbf{Operational Questions:} A set of 25 distinct operational questions (see Appendix \ref{appendix:operational_questions}) was created to assess the proficiency in retrieving specific factual information and performing analyses using the simulation output. These questions were designed to cover various aspects of the simulated operation, with an approximately uniform distribution across key entities and stages such as supplier interactions, worker activities, AGV and forklift utilization and package flow.

\textbf{Investigative Questions:} To specifically evaluate the  capabilities in identifying operational bottlenecks, three distinct investigative scenarios were simulated. Each scenario introduced a specific type of inefficiency into the baseline model, mirroring potential real-world disruptions:
\begin{itemize}

    \item \textbf{Scenario 1: Delay in Stage Transfer:} For a particular supplier, a specific process inefficiency was simulated, primarily introducing significant delays at a specific stage. This initial bottleneck led to significantly prolonged overall discharge times for their packages and created a downstream imbalance in equipment utilization.
    \item \textbf{Scenario 2: Degraded Forklift Performance:} One specific forklift was modeled to operate with reduced efficiency throughout its designated shift, leading to localized congestion and delays in tasks reliant on that particular forklift.
        \item \textbf{Scenario 3: Supplier-Specific Processing Delay:} For a particular supplier, targeted inefficiencies were simulated, introducing increased handling and suboptimal task allocation within the unloading and package processing stages, leading to significantly prolonged processing times.
\end{itemize}
For each of these three systematically perturbed scenarios, a unique investigative question was formulated. The objective of each such question was to task the framework with identifying the primary operational bottleneck or pinpointing the most significant performance degradation resulting from the deliberately introduced inefficiency.

\begin{table*}[!t]
\centering

\begin{flushleft}
\small
{\textbf{Note:} The LLM Agent formulates its own sequence of questions based on its reasoning framework and evolving evidence from the KG. Human expert iterative questions for this investigation: (1) What is the total discharge time for CamelCargo compared to the global average? (2) What are the equipment utilization rates for AGVs and forklifts and do they show an imbalance? (3) To find the bottleneck, what are the package-level waiting times at each distinct process stage? (4) Does any specific stage show a significant deviation from the global average and is this the primary cause of the delay and the equipment utilization imbalance? The LLM generations have been condensed to fit inside the table. Cypher queries are highly condensed conceptual representations for brevity.}
\end{flushleft}
\small
\begin{tabular}{
    >{\arraybackslash}p{0.23\textwidth} | 
    >{\arraybackslash}p{0.19\textwidth} | 
    >{\arraybackslash}p{0.21\textwidth} | 
    >{\arraybackslash}p{0.27\textwidth} 
}

\toprule 
\textbf{LLM Agent - Iterative Question} & \textbf{LLM Agent - Plan (Condensed)} & \textbf{KG ResultSet (Condensed)} & \textbf{Cypher Query (Condensed)}\\
\midrule
 What is the total unload time for supplier CamelCargo in the Discharge Flow and how does it compare to the global average unload time for all suppliers?
 &
Compare CamelCargo's total unload time against the global average for all suppliers, showing the difference and ratio.
&
Total Unload Time :

6,870s (1.39x the global average of 4,955s)
&  \textit{CALL \{MATCH (s:SUPPLIER \{supplier\_id: 'CamelCargo'\}) ..... gl\_avrg\_unload\_sec AS difference\_sec, toFloat(camel\_ul\_sec)/ gl\_avrg\_unload\_sec AS ratio\_to\_gl}
\\
\cmidrule(lr){1-4}

What is the supplier waiting time for CamelCargo in the Discharge Flow and how does it compare to the global average supplier waiting time for all suppliers?
 & 
Compare CamelCargo's waiting time against the global average waiting time for all suppliers.
. & 
Supplier CamelCargo Wait Time : 22.0s (vs. global average of 21.6s)
 & \textit{CALL \{MATCH (s:SUPPLIER \{supplier\_id: 'CamelCargo'\}) -[stw: SUPPLIER\_TO\_WORKER] -\textgreater(w:WORKER) ..... RETURN gl\_avrg\_wtng\_time\_sec}

 \\
 \cmidrule(lr){1-4}

What is the utilization rate of AGVs and forklifts during the discharge process for supplier CamelCargo and how does it compare to the global average utilization rate for these equipments?
& 
Calculate the AGV and forklift utilization rates for supplier CamelCargo and compare them against the global average rates.
 & 
Equipment Utilization -
AGV Utilization: 54.8\% (Global Avg: 41.7\%)
Forklift Utilization: 23.7\% (Global Avg: 40.1\%)
&  \textit{CALL \{MATCH (s:SUPPLIER
\{supplier\_id: 'CamelCargo'\})
-[stw: SUPPLIER\_TO\_WORKER] -\textgreater(w:WORKER)
...
CASE WHEN fl\_act\_gl \textgreater 0 THEN toFloat(fl\_p\_gl)
/ fl\_act\_gl ELSE 0 END AS fl\_util\_gl}
\\
\cmidrule(lr){1-4}

What are the package waiting times at each stage of the discharge process for supplier CamelCargo and how do they compare to the global average waiting times at each stage?
&
For each package from CamelCargo, show its waiting time at each process stage alongside the global average for that stage.

 & 
Package Wait Times (vs. Global Avg) -
Wait for Worker: Significantly Higher (e.g., one package waited 3,654s vs. the 1,375s avg)
Wait for AGV, Forklift, Storage: At or below average

 &  \textit{CALL \{MATCH (s:SUPPLIER)-
[stw: SUPPLIER\_TO\_WORKER]-\textgreater(w:WORKER)
...
WITH DISTINCT stw.package\_id AS package\_id,
s, stw, wta, atf, fts
...
gl\_avrg\_wait\_to\_storage\_sec
ORDER BY package\_id}
\\

\midrule 
\multicolumn{4}{p{\linewidth}}{%

\textbf{Agent Summary:} Supplier CamelCargo's discharge process was the longest, taking 6,870 seconds (1.39 times the global average), primarily due to extreme "Wait to Worker" times. Several packages waited over double the average time for a worker (e.g., one package waited 3,654 seconds vs. the 1,375-second average). This initial bottleneck created an imbalance downstream, with AGV utilization being higher than average (0.54 vs. 0.42) and forklift utilization being much lower (0.24 vs. 0.40).
} \\
\bottomrule
\end{tabular}
\caption{Scenario 1: Why did CamelCargo's discharge take longer than others?}
\label{tab:case_study_camelcargo}
\end{table*}

\section{Results and Discussion}
 For operational queries performance was measured using the pass@k metric\cite{chen2021evaluating} to assess answer accuracy across several attempts. This was benchmarked against two baselines: (i) single-pass Cypher generation with answer synthesis and (ii) an enhanced version adding post-answer self-reflection. For investigative scenarios, our iterative Reasoning chain refining each step based on accumulated evidence was evaluated by a human expert.

\subsection{Performance on Operational QA}
The experimental results for operational question answering presented in Table \ref{tab:operational_qa_results} highlight the significant advantages of our proposed approach. While incorporating a self-reflection (SR) mechanism into a direct question-answering pipeline (Baseline: Direct QA + SR) does offer a substantial improvement over a simple, single-pass baseline: Direct QA, our proposed method consistently outperforms both baselines, particularly in achieving comprehensive correctness as indicated by the maximum Pass@2 scores across all operational stages.

The results directly validate our hypothesis, contrasting the brittleness of the baseline's single-pass approach with the robustness of our guided, iterative framework. The baseline's low average Pass@1 (P@1) score of 0.59 confirms that its complex, monolithic queries are highly prone to failure, dropping as low as 0.33 for the "Worker" category. The improved baseline with self-reflection(SR) also shows minor improvement of 0.42 P@1 in this category. In sharp contrast, our framework achieves an average P@1 score of 0.92. This demonstrates that our method, which decomposes the query and embeds self-reflection at each step, is vastly more likely to produce a correct solution on its very first attempt

The Pass@2 (P@2) scores further underscore this robustness. Our framework achieves a perfect average P@2 of 1.00, meaning that when generating two independent solution paths, at least one of them was correct 100\% of the time. The baseline, however, only scores 0.71 P@2. This is a critical finding: even when the baseline is given two independent chances to generate a monolithic query, it still fails to produce a single correct answer 22\% of the time.

The combination of a 0.92 P@1 and a 1.00 P@2 strongly suggests that our agent's step-wise, reflective process consistently generates high-quality, correct solutions. In the rare 8\% of cases where its first independent attempt fails, its second independent attempt succeeds, proving the fundamental stability and reliability of our decompositional approach.
The ability to decompose, validate and refine sub-questions at each stage enhances the agent's capability in both direct question answering and complex diagnostic scenarios.

\subsection{Performance on Investigative QA}
We present three case studies to evaluate the agent’s effectiveness in handling investigative QA. Due to space constraints, only one is discussed in detail. Others can be found in Appendix \ref{additional scenarios}.  The CamelCargo investigation (scenario 1) represents a complex diagnostic scenario involving significant discharge delays that require systematic analysis across multiple operational stages. The investigation revealed (see Table \ref{tab:case_study_camelcargo}) that CamelCargo's total discharge time of 6,870 seconds significantly exceeded the global average discharge time of 4,955 seconds, representing a 39\% performance degradation (1.39 times the global average).
Expert analysis through the framework identified that the primary contributing factor was extreme inefficiencies in the worker stage, where packages experienced significantly prolonged waiting times before worker assignment. Specific instances revealed packages waiting 3,654 seconds compared to the global average of 1,375 seconds, creating a critical bottleneck that cascaded through the entire discharge process.

Further investigation of the CamelCargo scenario demonstrates sophisticated diagnostic capabilities of the workflow through a systematic sub-question generation and evidence-based analysis (see Table \ref{tab:case_study_camelcargo}.) 

\begin{table*}[!t]
\centering

\small
\setlength{\tabcolsep}{3pt} 

        \begin{tabular}
        {
        >{\centering\arraybackslash}p{0.11\textwidth} | 
    >{\centering\arraybackslash}p{0.11\textwidth} | 
    >{\centering\arraybackslash}p{0.11\textwidth} |
    >{\centering\arraybackslash}p{0.10\textwidth} | 
    >{\centering\arraybackslash}p{0.12\textwidth} | 
    >{\centering\arraybackslash}p{0.11\textwidth} |
    >{\centering\arraybackslash}p{0.11\textwidth} | 
    >{\centering\arraybackslash}p{0.10\textwidth} 
        }
        
\toprule
\textbf{Evaluation Metric} &   \textbf{Clarity} &
 \textbf{Logical Coherence}&
 \textbf{Necessity} &\textbf{Completeness} &\textbf{Factuality}& \textbf{Sufficiency
}&\textbf{Query Quality} \\

\midrule
Score (0-10)&  8.04 &
8.21&
9.0&
8.58&
8.46&
8.0&
7.96
 \\
\bottomrule
\end{tabular}
\small
\caption{Performance on Investigative Queries evaluated on various metrics}
\label{tab:investigative_results}
\end{table*}
\textbf{Initial Investigation Phase:} The investigation began with a comparative analysis: "What is the total unload time for supplier CamelCargo in the Discharge Flow and how does it compare to the global average unload time for all suppliers?" This revealed a significant issue, with CamelCargo's unload time of $6,870$ seconds exceeding the global average of $4,955$ seconds.

\textbf{Progressive Diagnostic Refinement:} The framework demonstrated sophisticated reasoning progression through strategically generated sub-questions that systematically explored potential bottleneck sources. The investigation progressed through multiple phases:
\begin{itemize}
    \item \textbf{Supplier Waiting Time Analysis:} Comparing CamelCargo's supplier waiting time of 22.0 seconds against the global average of 21.6 seconds, which appeared within normal ranges and ruled out supplier-side delays.
    \item \textbf{Equipment Utilization Assessment:} Analyzing AGV and forklift utilization rates revealed an imbalanced resource allocation pattern. AGV utilization was significantly higher than average (54.8\% vs. 41.7\% global average), while forklift utilization was substantially lower (23.7\% vs. 40.1\% global average), indicating downstream process inefficiencies.
    \item \textbf{Stage-by-Stage Package Analysis:} The framework systematically examined package waiting times at each discharge stage, revealing the critical bottleneck in the "Wait to Worker" phase.
\end{itemize}

\textbf{Evidence Integration and Causal Analysis:} The framework demonstrated advanced evidence integration capabilities by identifying that while waiting times for AGV, forklift and storage operations remained at or below global averages, the "Wait to Worker" stage showed extreme variability and delays. This created a cascading effect where the initial bottleneck led to resource imbalances downstream, explaining the higher AGV utilization and lower forklift utilization patterns.
The investigation conclusively identified that CamelCargo's performance issues stemmed from worker assignment inefficiencies rather than equipment-related bottlenecks, with specific packages experiencing wait times more than double the global average, fundamentally disrupting the entire discharge flow efficiency. A similar performance was observed against other two investigative scenarios as shown in Appendix \ref{additional scenarios}.

In addition to this, a comprehensive human evaluation of our framework across seven critical quality dimensions demonstrates exceptional performance, with scores ranging from 7.96 to 9.0 on a 10-point scale (see Table \ref{tab:investigative_results}). These results represent human expert assessment of over 12 investigation analysis logs produced by our intelligent assistant based on the 3 aforementioned scenarios, providing robust validation of diagnostic quality across diverse warehouse bottleneck scenarios. For more details refer to appendix \ref{additional_info_on_investigative_qa_eval}. This multi-dimensional assessment approach, inspired by recent advances in dialectical evaluation frameworks for LLM reasoning chains \cite{anghel2025pearl}, provides a holistic view of system performance that extends beyond traditional accuracy metrics to encompass the nuanced quality requirements essential for practical warehouse diagnostic applications.

\section{Implications and Limitations}

Our framework transforms warehouse planning by providing intuitive access to complex simulation insights across multiple operational horizons. The high pass@k scores (Table \ref{tab:operational_qa_results}) enable planners to obtain precise, real-time visibility into supplier interactions, resource utilization and package flows, supporting both daily operations and tactical adjustments without requiring specialized technical expertise.

Most significantly, the investigative capabilities move beyond surface-level reporting to deliver meaningful diagnostic insights. By systematically querying simulation-derived Knowledge Graphs through LLM-driven reasoning, the framework isolates root causes of performance issues and reveals subtle bottlenecks often missed by traditional analytics. This creates a more powerful and interpretable warehouse digital twin that enables targeted, data-driven interventions such as process redesign and resource reallocation, ultimately supporting more adaptive and informed warehouse planning.

Several limitations require consideration for operational deployment. The initial Knowledge Graph schema design demands substantial domain expertise and engineering investment, though this can be mitigated through standardized schema templates and automated generation tools for common warehouse configurations.

Despite robust self-correction capabilities, the absolute reliability of Cypher query generation and synthesized explanations requires ongoing validation, particularly for novel operational scenarios. This necessitates continuous monitoring procedures and confidence scoring mechanisms for critical diagnostic conclusions.

The framework's generalizability, validated primarily in warehouse unloading contexts, requires demonstration across broader warehouse processes and different simulation frameworks. Future work must systematically evaluate performance across picking, packing, inventory management and other operational contexts.

\section{Conclusion and Future Work}

Extracting actionable insights from voluminous DES data is a significant challenge for timely warehouse decision-making. To solve this, we propose a novel framework integrating KGs with a reasoning-capable LLM agent, providing a more intuitive and powerful way to interact with simulation data. The architecture combines a QA chain with step-wise guidance and an iterative reasoning chain equipped with sub-questioning, Cypher query generation and self-reflection. This enables both high-accuracy operational queries and deeper, evidence-driven investigations into system inefficiencies. Experimental evaluations demonstrate the framework's proficiency in accurately answering operational questions and, more significantly, its robust capability in performing iterative, evidence-driven investigations to identify operational bottlenecks in simulated scenarios, surpassing traditional baseline methods.\\
In the future, we plan to integrate advanced reasoning LLM architectures to enhance the agent's diagnostic depth and efficiency. To validate robustness, we will expand the framework's application beyond unloading to a wider array of warehouse operations (e.g., slotting, picking, loading, inventory management), integrating both simulated and on-field data. This expansion necessitates developing rigorous benchmarking methodologies to formally quantify performance. Ultimately, this work is a stepping stone toward developing a fully autonomous industrial planning and scheduling agent.

\section*{Acknowledgments}
This work is supported by PhiLabs, Quantiphi Inc. We would like to thank Dr. Dagnachew Birru and Mr. Asif Hasan for their continued support.

 \small

\appendix
\onecolumn
\section{Appendix}
\subsection{Warehouse Resources}
\label{appendix:warehouse_details}
The Table \ref{tab:resource_information} highlights the various resource types and their respective ids that are modeled in the simulation.
\begin{table}[ht]
\centering

\begin{tabular}{
 |>{\arraybackslash}p{0.2\textwidth} | 
    >{\arraybackslash}p{0.6\textwidth} |}
\hline
Resource Type & Resource IDs                                                                                                                                                                                                                       \\ 
\hline
Supplier      &  AuroraFarms, BlackSheepDist, CamelCargo, DeltaDrops, EvergreenEdge                                                                                                                      \\ 
\hline
Worker & BW\_02, BW\_00, BW\_01, BW\_03, BW\_09,BW\_10, BW\_11, BW\_08, BW\_05, BW\_07, BW\_06, BW\_04                                                                                      \\ 
\hline
AGV           & AGV\_10, AGV\_12, AGV\_11, AGV\_00, AGV\_01, AGV\_02, AGV\_03, AGV\_13,AGV\_14, AGV\_04, AGV\_15, AGV\_05, AGV\_16, AGV\_08, AGV\_17, AGV\_09, AGV\_07, AGV\_18, AGV\_19, AGV\_06 \\ 
\hline
Fork Lift     & FL\_00, FL\_01, FL\_04, FL\_02, FL\_03                                                       \\ 
\hline
Storage Block & A, B, C, D                \\
\hline
\end{tabular}
\caption{The resource ids for the different resource present in the simulated scenario}
\label{tab:resource_information}
\end{table}

\

\subsection{KG Schema}
\label{appendix:KGSchema}
\textbf{Node Properties:}
\begin{itemize}
    \item \textbf{SUPPLIER:}\texttt{ supplier\_id: STRING, arrival\_time: DATETIME, discharge\_start:DATETIME, discharge\_end: DATETIME}

    \item \textbf{WORKER:} \texttt{ worker\_id: STRING}
    \item \textbf{AGV:} \texttt{ agv\_id: STRING}
    \item \textbf{FL (Forklift):}\texttt{ forklift\_id: STRING}
    \item \textbf{STORAGE:} \texttt{ block\_id: STRING}
\end{itemize}

\textbf{Relationship Properties:}
\begin{itemize}
\item \textbf{SUPPLIER\_TO\_WORKER} \\ \texttt{package\_id:STRING, worker\_pick\_up\_start:DATETIME}
\item \textbf{WORKER\_TO\_AGV} \\
\texttt{package\_id: STRING, agv\_arrival:DATETIME, agv\_journey\_start:DATETIME, worker\_pick\_up\_end: DATETIME}
\item \textbf{AGV\_TO\_FL} \\ \texttt{package\_id: STRING, agv\_journey\_end: DATETIME, fl\_placement\_start: DATETIME}
\item \textbf{FL\_TO\_STORAGE} \\ \texttt{package\_id: STRING, fl\_placement\_end: DATETIME}
\end{itemize}

\subsection{Multi-Dimensional Quality Assessment for Investigative QA}
\label{additional_info_on_investigative_qa_eval}
The human evaluation results reveal consistently high performance across all measured dimensions, with Necessity achieving the highest score of 9.0, indicating that human evaluators consistently found the framework's investigation paths to be highly relevant and essential for comprehensive bottleneck diagnosis. This exceptional necessity score, validated through human expert judgment across multiple investigation scenarios, confirms the effectiveness of our structured sub-question generation methodology in identifying and pursuing only the most critical investigative directions while avoiding redundant or irrelevant analysis paths.

Logical Coherence (8.21) and Clarity (8.04) scores demonstrate that human evaluators consistently found the framework's reasoning flow to be systematic and the diagnostic explanations to be easily interpretable. These high human-assessed scores reflect the success of our Prompt Refinement and Sequencing architecture in guiding LLM reasoning toward structured, logical investigation patterns that effectively communicate with warehouse planning professionals. The coherence score particularly validates our sequential reasoning methodology's ability to maintain logical consistency across complex multi-stage investigations as perceived by human domain experts.

Completeness (8.58) and Factuality (8.46) scores indicate that human evaluators consistently found robust coverage of relevant diagnostic dimensions and high accuracy in data interpretation and evidence synthesis across the analyzed investigation logs. The strong human-assessed completeness score demonstrates that our investigative analysis engine successfully addresses all critical aspects of bottleneck scenarios as judged by expert evaluators, while the high factuality score validates the effectiveness of our dynamic few-shot retrieval system and enhanced Knowledge Graph integration in supporting accurate diagnostic conclusions that align with human expert expectations.

Sufficiency (8.0) reflects human evaluator assessment of adequate investigation depth across diverse operational scenarios, while Query Quality (7.96) indicates human judgment of generally excellent Cypher query generation and execution efficiency. The slightly lower query quality score, while still representing strong performance in human evaluation, suggests potential areas for continued optimization in automated query generation algorithms, particularly for highly complex multi-stage investigative scenarios as perceived by human domain experts.

The significance of these human-evaluated results cannot be overstated, as they represent validation by domain experts who can assess not only technical correctness but also practical utility and interpretability for real-world warehouse planning applications. These results align with findings from dialectical evaluation research \cite{anghel2025pearl}, which emphasizes that multi-dimensional assessment frameworks provide more comprehensive and actionable insights than traditional single-metric evaluations. The consistently high human evaluation scores across all dimensions validate our framework's advancement over baseline approaches and demonstrate readiness for practical deployment in warehouse planning applications where diagnostic reliability and explanation quality are critical success factors as assessed by human practitioners.
The strong human evaluation performance across both technical dimensions (Query Quality, Factuality) and communicative dimensions (Clarity, Coherence) indicates successful integration of sophisticated AI capabilities with practical usability requirements that meet human expert standards, supporting our framework's potential for widespread adoption in industrial warehouse management contexts where human decision-makers must rely on and trust the diagnostic insights provided by the system.

\subsection {Additional Scenarios}
\label{additional scenarios}

\begin{table*}[!ht]

\begin{flushleft}
{\textbf{Note:} The LLM Agent formulates its own sequence of questions. Potential human expert iterative questions for a forklift delay investigation might include: (1) KPIs per forklift (tasks, durations, idle times)? (2) Which forklifts show consistently longer task/idle times? (3) Operational time breakdown for suspect forklifts? (4) Queue/wait times for packages per forklift? (5) Are delays linked to specific zones/tasks/times for suspect forklifts? (6) How does a suspect forklift's throughput compare? (7) Any error codes/issues reported? Cypher queries are highly condensed conceptual representations.}
\end{flushleft}

\begin{tabular}{
    >{\arraybackslash}p{0.18\textwidth} | 
    >{\arraybackslash}p{0.16\textwidth} | 
    >{\arraybackslash}p{0.18\textwidth} | 
    >{\arraybackslash}p{0.38\textwidth} 
}

\toprule 
\textbf{LLM Agent - Iterative Question} & \textbf{LLM Agent - Plan (Condensed)} & \textbf{KG ResultSet (Condensed)} & \textbf{Cypher Query (Condensed)}\\
\midrule
What is the average waiting time for each forklift during the discharge flow and how does it compare to the global average waiting time for all forklifts?
&
Compare the average waiting time for each forklift (FL\_00-FL\_04) against the global average forklift waiting time.
&
Avg Wait (vs. 221.33s Global):
FL\_00: 363.78s (High)
FL\_04: 225.91s (Slightly High)
FL\_01: 205.91s (Low)
FL\_03: 182.89s (Low)
FL\_02: 165.41s (Low)
&  \textit{CALL \{MATCH (fl:FL)... WITH fl, collect(\{start: fl\_start, end: fl\_end\}) AS segs... RETURN fl.forklift\_id, ... avg\_wait\_seconds\} 
CALL \{...reduce(all\_waits = [], w IN waits\_by\_fl | all\_waits + w) AS all\_waits... RETURN ... global\_avg\_wait\_seconds\} 
RETURN forklift\_id, ... global\_avg\_wait\_seconds}
\\
\cmidrule(lr){1-4}
What is the utilization percentage for each forklift during the discharge flow, specifically for Forklift FL\_00, FL\_01, FL\_02, FL\_03 and FL\_04?
& 
Calculate and list the utilization percentage for each forklift from FL\_00 to FL\_04.
& 
Forklift Utilization:
FL\_00: 30.26\%
FL\_01: 36.31\%
FL\_02: 42.45\%
FL\_03: 40.16\%
FL\_04: 35.05\%
& \textit{MATCH (fl:FL) WHERE fl.forklift\_id IN [...] ... WITH fl.forklift\_id AS forklift\_id, sum(...) AS total\_handling\_seconds, min(fl\_start) AS first\_start, max(fl\_end) AS last\_end ... RETURN forklift\_id, CASE WHEN span\_seconds \textgreater 0 THEN toFloat(...) / span\_seconds * 100.0 ELSE null END AS utilization\_percentage}
\\

\midrule 
\multicolumn{4}{p{\linewidth}}{%
\textbf{Agent Summary:} \textbf{The differences in forklift waiting times during the discharge flow reveal variations in efficiency and utilization among the forklifts.} Forklift FL\_00 has the highest average waiting time of 363.78 seconds, which is significantly higher than the global average waiting time of 221.33 seconds. This suggests potential inefficiencies or bottlenecks in its operation. In contrast, forklifts FL\_01, FL\_02 and FL\_03 have average waiting times of 205.91 seconds, 165.41 seconds and 182.89 seconds respectively, all of which are lower than the global average, indicating more efficient operations. Forklift FL\_04 has a slightly higher waiting time of 225.91 seconds compared to the global average, suggesting a minor inefficiency. Forklift FL\_00, with the highest waiting time, has the lowest utilization at 30.26\%. Conversely, FL\_02, which has the lowest waiting time, shows the highest utilization at 42.45\%. Forklifts FL\_01, FL\_03 and FL\_04 have utilizations of 36.31\%, 40.16\% and 35.05\% respectively, reflecting their moderate efficiency levels. These differences in waiting times and utilization shows the varying levels of operational efficiency among the forklifts during the discharge flow.} \\
\bottomrule
\end{tabular}
\caption{Scenario 2: What do the differences in forklift waiting times reveal about the discharge flow?}
\label{tab:case_study_forklift_wait}
\end{table*}

\begin{table*}[!ht]

\begin{flushleft}
{\textbf{Note:} Human expert iterative questions for this investigation: (1) What is the total discharge time for AuroraFarms and how does it compare to the global average? (2) What are the average package waiting times and worker operation times for AuroraFarms compared to the global average? (3) What are the equipment utilization rates (AGV, forklift) for AuroraFarms and how do they compare to the global averages? (4) What is the average number of active AGVs and forklifts servicing AuroraFarms compared to the global average? (5) Are there specific bottlenecks in resource deployment or management (e.g., low utilization despite high equipment availability) causing the delay? The LLM generations have been condensed to fit inside the table. Cypher queries are highly condensed conceptual representations for brevity.}

\end{flushleft}
\small

 \begin{tabular}{
    >{\arraybackslash}p{0.23\textwidth} | 
    >{\arraybackslash}p{0.18\textwidth} | 
    >{\arraybackslash}p{0.13\textwidth} | 
    >{\arraybackslash}p{0.38\textwidth} 
}

\toprule 
\textbf{LLM Agent - Iterative Question} & \textbf{LLM Agent - Plan (Condensed)} & \textbf{KG ResultSet (Condensed)} & \textbf{Cypher Query (Condensed)}\\
\midrule
What is the average total unload time for the supplier AuroraFarms compared to the global average across all suppliers in the discharge flow?
&
Compare AuroraFarms' average unload time against the global average unload time.
&
AuroraFarms: 8,923s
Global Avg: 6,724.6s
&  \textit{CALL \{MATCH (s:SUPPLIER \{supplier\_id: 'AuroraFarms'\})... RETURN AVG(...) AS aurora\_avg\_seconds\} 
CALL \{MATCH (s:SUPPLIER) ... RETURN AVG(...) AS global\_avg\_seconds\} 
RETURN aurora\_avg\_seconds, global\_avg\_seconds}

\\
\cmidrule(lr){1-4}

What is the average package waiting time at the waiting point for the supplier AuroraFarms compared to the global average in the discharge flow?
& 
Compare AuroraFarms' package wait time at the waiting point against the global average.
& 
AuroraFarms: 250.26s
Global Avg: 338.21s
(Difference: -87.94s)
& \textit{CALL \{MATCH (s:SUPPLIER \{supplier\_id: 'AuroraFarms'\})...wta... RETURN avg(per\_pkg\_wait\_secs) AS supplier\_avg\_wait\_seconds\} 
CALL \{MATCH (s:SUPPLIER)...wta... RETURN avg(per\_pkg\_wait\_secs) AS global\_avg\_wait\_seconds\} 
RETURN ... difference\_seconds}
 
 \\
 \cmidrule(lr){1-4}

What is the average time taken for workers to pick up packages from the supplier AuroraFarms and carry them to the waiting point compared to the global average in the discharge flow?
& 
Compare AuroraFarms' worker pickup/carry time against the global average.
& 
AuroraFarms: 58.24s
Global Avg: 58.18s
& \textit{ CALL \{...MATCH (s:SUPPLIER \{supplier\_id: 'AuroraFarms'\})...stw...wta... RETURN avg(...) AS supplier\_avg\_seconds\} 
CALL \{...MATCH (s:SUPPLIER)...stw...wta... RETURN avg(...) AS global\_avg\_seconds\} 
RETURN supplier\_avg\_seconds, global\_avg\_seconds}
\\
\cmidrule(lr){1-4}

What is the utilization rate of AGVs servicing the supplier AuroraFarms compared to the global average utilization rate of AGVs in the discharge flow?
&
Compare AuroraFarms' AGV utilization rate against the global average.
& 
AuroraFarms AGV Util: 0.1217 (12.2\%)
Global Avg AGV Util: 0.4170 (41.7\%)
& \textit{ CALL \{MATCH (s:SUPPLIER)...(a:AGV)... WITH agv\_id, supplier\_id, ... CASE WHEN supplier\_id = 'AuroraFarms' THEN ... END ...\} 
WITH collect(...) AS rows 
... RETURN aurora\_agv\_avg\_utilization, global\_agv\_avg\_utilization}
\\
\cmidrule(lr){1-4}

What is the utilization rate of forklifts servicing the supplier AuroraFarms compared to the global average utilization rate of forklifts in the discharge flow?
&
Compare AuroraFarms' forklift utilization rate against the global average.
&
AuroraFarms FL Util: 0.0678 (6.8\%)
Global Avg FL Util: 0.3321 (33.2\%)
&
\textit{MATCH (s:SUPPLIER)...(fl:FL)... 
WITH fl, s, ... sum(CASE WHEN s.supplier\_id = 'AuroraFarms' THEN ... END) AS aurora\_seconds 
... RETURN AVG(aurora\_util) AS auroraFarms\_utilization\_rate, AVG(global\_util) AS global\_average\_utilization\_rate}
\\
\cmidrule(lr){1-4}

What is the average number of AGVs and forklifts actively servicing the supplier AuroraFarms compared to the global average number of AGVs and forklifts servicing suppliers in the discharge flow?
&
Compare the average number of active AGVs and forklifts for AuroraFarms against the global average.
&
Avg AGVs: 4.32 (Global: 4.25)
Avg Forklifts: 3.04 (Global: 2.59)
&
\textit{CALL \{MATCH (s:SUPPLIER \{supplier\_id: 'AuroraFarms'\})... RETURN ... aurora\_avg\_agvs, aurora\_avg\_fls\} 
... CALL \{...\} ...
CALL \{...\} ... 
RETURN aurora\_avg\_agvs, aurora\_avg\_fls, global\_avg\_agvs, global\_avg\_fls }
\\

\midrule 
\multicolumn{4}{p{\linewidth}}{%
\textbf{Agent Summary:} \textbf{The supplier AuroraFarms experienced slower discharge times primarily due to a significantly higher average total unload time compared to the global average.} The average total unload time for AuroraFarms was 8,923 seconds, which is considerably longer than the global average of 6,724.6 seconds.
Despite having a shorter average package waiting time at the waiting point (250.26 seconds) compared to the global average (338.21 seconds) and a similar average time for workers to pick up packages (58.24 seconds compared to the global average of 58.18 seconds), the overall discharge process was slower. A critical factor contributing to the slower discharge was the low utilization rates of both AGVs and forklifts servicing AuroraFarms. The utilization rate of AGVs was 0.1217, much lower than the global average of 0.4107 and the utilization rate of forklifts was 0.0678, compared to the global average of 0.3321. Although the average number of AGVs and forklifts actively servicing AuroraFarms was slightly higher than the global average, this did not compensate for the low utilization rates, indicating inefficiencies in resource deployment or management.
} \\
\bottomrule
\end{tabular}
\caption{Scenario 3: Why was the supplier AuroraFarms discharge slower than others?}
\label{tab:case_study_aurorafarms}
\end{table*}


\label{appendix:operational_questions}
\begin{table*}[ht]
\centering

\begin{tabular}{|l|p{14cm}|} 
\hline
\textbf{Category} & \textbf{Question} \\ 
\hline
SUPPLIER & What is the number of discharge processes that are completed on an hourly basis? 
\\ 
 & Where and how many containers discharged from supplier DeltaDrops distributed in each block in the storage?              \\ 
 & Which supplier had the shortest total discharge time and how many packages were moved?                                   \\ 
 & What is the average waiting time for a supplier truck before unloading begins? Which truck waited the most?              \\ 
 & Which hour had the most total waiting time during package unload?                                                       \\ 
\hline

WORKER   & For each person, what was the total number of packages they handled during a shift?                                      \\ 
   & What is the average time taken by a person to move a package from truck to  AGV? Who is the most efficient person ?  \\ 
   & How much time does each worker take to unload all packages from supplier DeltaDrops?                                     \\ 
   & How many workers were used to unload packages from supplier CamelCargo?                                     \\ 
   & Which workers were assigned to most number of suppliers?     \\                    \hline                            
AGV      & Which three AGVs processed the least amount of packages?                                                                 \\ 
      & What is the average travel time for an AGV to move a package from the dock to its assigned storage area?                 \\ 
      & How many trips does each agv make during unloading along with the average journey time?                                  \\ 
& How many packages did AGV 04 handle from each supplier ? \\
& Which AGV was the least utilized ?\\

\hline

FORKLIFT & Which package waited the longest for a fork lift ?                                                                       \\ 
 & How many packages are handled by each forklift?                                                                          \\ 
 & Which forklift is the most under utilized ?                                                                              \\ 
 & What is the average time taken by a forklift to move a package to its assigned storage space?                            \\ 
 & What is the utilization rate (percentage of time in use) for each forklift?                                              \\ 
\hline
PACKAGE  & which storage block contains the highest number of containers?                                                           \\ 
  & What is the average time a package discharge takes?                                                                      \\ 
  & What is the average waiting time for a package to be transferred to a forklift after AGV arrival at the storage area?    \\ 
  & Which package experienced the longest total time from arrival at the dock to placement in its final storage location?    \\ 

  & How many packages took longer than the average unload time during and what is the average discharge time?                \\ 
  & Which packages were handled by both AGV 10 and forklift 00?                                                              \\ 
\hline
\end{tabular}
\caption{Sample Operational Questions}
\label{tab:operational_questions}
\end{table*}

\end{document}